\documentclass[10pt,twocolumn,letterpaper]{article}

\usepackage[pagenumbers]{cvpr} % To force page numbers, e.g. for an arXiv version
\usepackage[accsupp]{axessibility}  % Improves PDF readability for those with disabilities.

\usepackage{multirow}
\usepackage{graphicx}
\usepackage{amssymb}
\usepackage{lineno}

\usepackage{xcolor}

\newcommand{\prob}{\mathrm{p}}
\newcommand{\Prob}{\mathrm{P}}

\newcommand{\video}{\mathbf{x}}
\newcommand{\depth}{\mathbf{d}}
\newcommand{\I}{\mathrm{I}}
\newcommand{\orrd}{\raisebox{-0.1em}{\includegraphics[width=3em,height=0.8em]{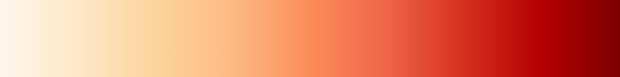}}}
\newcommand{\spectral}{\raisebox{-0.1em}{\includegraphics[width=3em,height=0.8em]{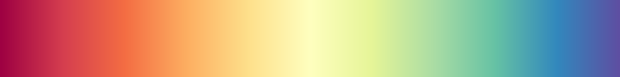}}}
\newcommand{\img}{\raisebox{-0.2em}{\includegraphics[width=1.4em,height=1.0em]{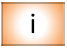}}}

\newcommand{\depthSnippet}{\raisebox{-0.2em}{\includegraphics[width=4.5em,height=1.0em]{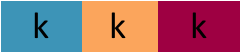}}}

\newcommand{\method}{RollingDepth}

\usepackage{arydshln} % For table dashed lines

\definecolor{cvprblue}{rgb}{0.21,0.49,0.74}
\usepackage[pagebackref,breaklinks,colorlinks,allcolors=cvprblue]{hyperref}
\usepackage{enumitem}

\def\paperID{707} %
\def\confName{CVPR}
\def\confYear{2025}

\title{Video Depth without Video Models}

\newcommand{\ethz}[0]{\textsuperscript{1}}
\newcommand{\cmu}[0]{\textsuperscript{2}}
\author{
Bingxin Ke\ethz \quad
Dominik Narnhofer\ethz\quad
Shengyu Huang\ethz \quad
Lei Ke\cmu \quad
Torben Peters\ethz \quad \\
Katerina Fragkiadaki\cmu \quad
Anton Obukhov\ethz \quad
Konrad Schindler\ethz \\
\vspace{2mm}
\ethz ETH Zurich \quad  \cmu Carnegie Mellon University
}

\begin{document}

\twocolumn[{%
\renewcommand\twocolumn[1][]{#1}%
\maketitle
%\begin{center}
\centering
  \includegraphics[width=\textwidth]{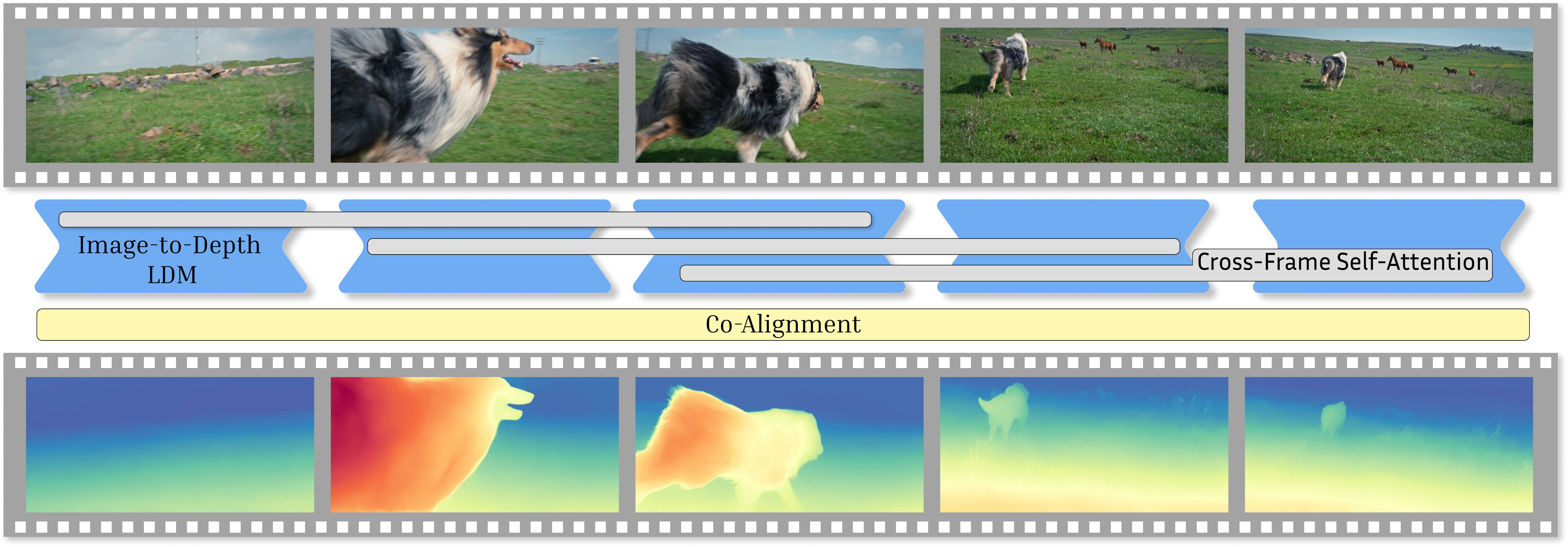}
  \vspace{-1.8em}
  \captionsetup{type=figure}
  \captionof{figure}{
      \textbf{The \method{} model} takes an unconstrained video and reconstructs a corresponding depth video. Unlike methods that rely on video diffusion models, it extends a single-image monodepth estimator such that it can process short snippets. To account for temporal context, snippets with varying frame rates are sampled from the video, processed, and reassembled through a global alignment algorithm to obtain long, temporally coherent depth videos. Depth is colour-coded near~\spectral~far.
  }
  \label{fig:teaser}
 %\end{center}%
 \vspace{1.8em}
 }]

% \maketitle
%%%%%%%%%%%%%%%%%%%%%%%% Abstract %%%%%%%%%%%%%%%%%%%%%%%%
\begin{abstract}
Video depth estimation lifts monocular video clips to 3D by inferring dense depth at every frame. Recent advances in single-image depth estimation, brought about by the rise of large foundation models and the use of synthetic training data, have fueled a renewed interest in video depth.
However, naively applying a single-image depth estimator to every frame of a video disregards temporal continuity, which not only leads to flickering but may also break when camera motion causes sudden changes in depth range.
An obvious and principled solution would be to build on top of video foundation models, but these come with their own limitations; including expensive training and inference, imperfect 3D consistency, and stitching routines for the fixed-length (short) outputs.
We take a step back and demonstrate how to turn a single-image latent diffusion model (LDM) into a state-of-the-art video depth estimator.
Our model, which we call \textbf{\method{}}, has two main ingredients: (i) a multi-frame depth estimator that is derived from a single-image LDM and maps very short video snippets (typically frame triplets) to depth snippets.
(ii) a robust, optimization-based registration algorithm that optimally assembles depth snippets sampled at various different frame rates back into a consistent video.
\method{} is able to efficiently handle long videos with hundreds of frames and delivers more accurate depth videos than both dedicated video depth estimators and high-performing single-frame models. 
Project page: \href{https://rollingdepth.github.io/}{rollingdepth.github.io}.
\end{abstract}

%%%%%%%%%%%%%%%%%%%%%%%% Intro %%%%%%%%%%%%%%%%%%%%%%%%
\section{Introduction}
\label{sec:intro}

% The task and its fundamental relevance
Inferring 3D scene structure from a video stream is a fundamental capability of a vision system. Besides its scientific relevance as an elementary building block of machine perception, it has a broad range of applications, including mobile robotics and autonomous driving, augmented reality, media production, and content creation.

% What is known but doesn't (always) work
Traditionally, a video would be converted into a 3D world model by recovering the camera trajectory with structure-from-motion (SfM) techniques~\cite{schonberger2016structure,griwodz2021alicevision}, then applying multi-view reconstruction based on either stereo triangulation~\cite{furukawa2015multi,yao2018mvsnet} or, more recently, inverse volume rendering~\cite{mildenhall2021nerf,kerbl2023_3d}. 
That approach has the attractive property that it delivers a full 3D scene model in a common coordinate frame. The price to pay is that it is only feasible under narrowly defined conditions: the camera motion must be just right, and the scene must have a static background with cooperative texture and lighting conditions.
In practice, both SfM and multi-view reconstruction fail more often than not when applied to in-the-wild videos.

% the alternative solution: monodepth
This is where video depth comes in. Not all applications require full-scale 3D reconstruction, and it turns out that information about the scene structure can be recovered much more reliably if one aims for a more modest goal: augment every video frame with a dense 2.5D depth map, in such a way that those depth maps are consistent through time.
The past years have witnessed tremendous progress in depth estimation from a single image, sidestepping camera pose estimation (and often also calibration of the focal length)~\cite{Ranftl2020_midas,ke2023repurposing,yang2024depthv2,bochkovskii2024prodepth}.
A common thread is that recent methods build on foundation models trained on internet-scale data, such as DINOv2~\cite{oquab2023dinov2} or StableDiffusion~\cite{rombach2022high}, and fine-tune them for depth estimation, often using predominantly synthetic RGB+depth image pairs that can be generated in large quantities and have accurate depth. The underlying, rich visual priors afford these depth estimators excellent zero-shot generalization across scene types, imaging, and lighting conditions.

% from single images to video
In general, applying a single-image depth estimator to a video frame-by-frame does not yield satisfactory results, but leads to depth flicker and drift.
These artifacts are caused by multiple factors.
Most obviously, neither the model training nor the inference procedure have any notion of temporal coherence between adjacent frames.
Moreover, monodepth estimation requires scene understanding, which may also suffer from the lack of temporal context (e.g., when a partially visible object only becomes recognizable after zooming out).
What is more, in a video the depth range between nearby and distant scene parts may change all of a sudden (e.g., when a foreground object enters the viewfield, or when the camera pans to a window), making consistent monodepth estimation difficult.

% video models
Some authors~\cite{shao2024learning, hu2024depthcrafter} have explored the idea of repurposing generative video models like Stable Video Diffusion~\cite{blattmann2023stable} for depth prediction.
These methods enable information exchange along the time axis and acquire a strong flow and motion prior during training, hence they achieve excellent local consistency through time.
On the downside, video LDMs -- besides being computationally demanding -- are trained for fixed, short sequence lengths and cannot be applied directly to uncurated footage of varying lengths.
To be practically useful, the diffusion routine must be wrapped into a partitioning scheme that splits the input video for processing and stitches the depth estimates back together, often resulting in low-frequency flickering and gradual drift.
We also find that current LDM-based video depth models tend to be less accurate on distant scene parts.

Rather than design more refined video LDMs, which require huge resources for training, we take a step back and re-examine how far one can take video depth estimation with augmented \emph{single-image} LDMs.
We design a set of measures that, taken together, extend a per-image monodepth framework like Marigold~\cite{ke2023repurposing} in a way that enables it to handle video input. Importantly, these measures greatly improve local and global consistency across time while maintaining a constant memory footprint such that one can process long sequences.
Specifically, we employ a “rolling” inference with a sliding window of a few frames (typically three, but other numbers are possible). Those snippets are sampled from the video with varying spacing, i.e., they can be immediately adjacent but also dilated along the timeline to cover long-range context. They are then fed into a multi-frame LDM fine-tuned from a single-frame model, with a modified cross-frame self-attention mechanism to enable information exchange.
To reassemble the snippets, we propose a robust optimization-based global co-alignment, followed by averaging the aligned frames.
Optionally, the resulting video can be degraded with moderate random noise and denoised again with the same per-snippet LDM to further refine spatial details.

To summarize, our approach estimates accurate and temporally consistent video depth without resorting to cumbersome video diffusion models. To that end, we contribute:
\begin{enumerate}
\item an LDM for monocular depth estimation in video snippets of a few frames, adapted from the Marigold~\cite{ke2023repurposing} single-frame model but able to capture temporal patterns across frames via self-attention; 
\item a rolling inference scheme that operates on snippets with multiple different (temporal) resolutions and enables efficient propagation of contextual information through video sequences of arbitrary length (up to minutes);
\item a global alignment procedure, based on robust optimization, to recompose the snippets into a depth video whose depth values remain consistent over long time periods;
\item an optional refinement of the final output with another round of multi-frame diffusion, where the same LDM is applied starting from a moderately degraded video.
\end{enumerate}

%%%%%%%%%%%%%%%%%%%%%%%% Related Work %%%%%%%%%%%%%%%%%%%%%%%%

\section{Related Work}
\label{sec:realated}

\subsection{Monocular Depth Estimation}
Monocular depth estimation is a dense regression task. The pioneering work by \citet{eigen_depth_2014} showed that metric depth values can be recovered from single sensors. 
Successive advancements include including 
various parameterizations (ordinals, bins, planar guidance maps, piecewise planarity, CRFs, \etc)~\cite{fu2018dorn, lee2019big, yuan2022newcrfs, patil2022p3depth, liu2023vadepth, Farooq_Bhat_2021, li2022binsformer, ning2023ait, Zhao_2022_CVPR}, switching CNN backbones to vision transformers~\cite{yang2021transformers, li2023depthformer, aich2021bidirectional,bhat2023zoedepth}, considering camera intrinsics~\cite{yin2023metric3d, hu2024metric3d, guizilini2023towards_zero_depth, piccinelli2024unidepth, piccinelli2025unidepthv2}, and patch-wise processing~\cite{li2023patchfusion, li2024patchrefiner, bochkovskii2024prodepth}.
To handle ``in-the-wild" settings, extensive internet photo collections are used for training~\cite{MDLi18_megadepth, yin2020diversedepth}.
MiDaS~\cite{Ranftl2020_midas} improves the generality by training on a mixture of multiple datasets.
Depth Anything~\cite{yang2024depthv1, yang2024depthv2} takes data scaling to the next level by relying on DINOv2~\cite{oquab2023dinov2}, a foundational model trained on 142M images in a self-supervised manner, and subsequently training with 62M pseudo-labels, 1M real depth annotations, and 0.5M synthetic ones.
Recent trends leverage generative models, particularly diffusion models~\cite{song2020denoising_ddim, ho2020denoising_ddpm}, for depth estimation~\cite{zhao2023vpd,duan2023diffusiondepth,saxena2023depthgen,zhao2023vpd}. Marigold~\cite{ke2023repurposing} proposed to fine-tune Stable Diffusion~\cite{rombach2022high}, a generative text-to-image latent diffusion model (LDM) trained with LAION-5B~\cite{schuhmann2022laion5b}, towards affine-invariant depth using 74k 
samples. This approach has been improved in many aspects including fewer steps~\cite{gui2024depthfm, garcia2024e2e, xu2024genpercept, he2024lotus}, finer details~\cite{zhang2024betterdepth}, and more modalities~\cite{fu2024geowizard, he2024lotus}.

\subsection{Video Depth Estimation}

Video depth estimation calls for dedicated mechanisms to ensure smoothness of adjacent frames, and correct handling of varying depth range. Existing approaches can be grouped into three main categories: test-time optimization, feed-forward prediction, and diffusion-based. \textit{Test-time optimization} methods~\cite{Luo-VideoDepth-2020, kopf2021rcvd, zhang2021consistent, chen2019self} often rely on camera poses or optical flow and perform optimization for each new video during inference. While these methods can produce depth estimates that are temporally consistent, their dependence on camera poses and long processing time hamper their application to open-world video scenarios. \textit{Feed-forward} prediction methods estimate depth sequences directly from input videos~\cite{li2023temporally, teed2018deepv2d, zhang2019exploiting, wang2022less,mamo, yasarla2024futuredepth}.
For example, DeepV2D~\cite{teed2018deepv2d} integrates camera motion estimation with depth prediction, MAMO~\cite{mamo} adopts memory attention mechanisms, and NVDS~\cite{NVDS, NVDSPLUS} introduces a stabilization network as a post-processing module. However, the generalization of these methods to in-the-wild videos is often constrained by the limited diversity of training data and model capacity. 

Very recently, concurrent with our work, several authors have investigated the use of video diffusion models, in particular SVD~\cite{blattmann2023stable}, for video depth. ChronoDepth~\cite{shao2024learning} DepthCrafter~\cite{hu2024depthcrafter} and DepthAnyVideo~\cite{yang2024depth} all modify video diffusion for conditional generative depth prediction. From the underlying video diffusion model they inherit high training and inference costs, and a restriction to short video clips of at most $\approx$100 frames. In contrast, in \method~we explore how to turn an image diffusion model into a temporally consistent depth estimator, which can handle long videos of 1000 frames or more.

\subsection{Image Diffusion Models for Video Tasks}
Image diffusion models have been employed in various video inverse problems, such as video generation, inpainting, and super-resolution~\cite{kwon2024solving, daras2024warped, Zhao_2023_ICCV_DDFM}. A large amount of work~\cite{qi2023fatezero, yang2023rerender,zhang2024towards} focusses on video editing, either by fine-tuning text-to-image diffusion models on video data~\cite{wu2023tune, liu2024video} or through training-free approaches using cross-frame attention and latent fusion~\cite{ceylan2023pix2video, khachatryan2023text2video}.
However, these works~\cite{wu2023tune, liu2024video,ceylan2023pix2video} predominantly address video-to-video translation tasks, where both the input and output reside in RGB space. In contrast, our approach leverages image diffusion priors to generate consistent depth videos, with the additional challenge to accommodate large variations of the depth range, as the near and far planes change -- often suddenly -- due to camera and object motion.
Implementation tricks when using single-image models, like fixing the initial noise or blending consecutive latent representations, can somewhat mitigate the lack of knowledge \wrt temporal coherence, but do not solve it~\cite{ke2023repurposing}.

%%%%%%%%%%%%%%%%%%%%%%%% Method %%%%%%%%%%%%%%%%%%%%%%%%

\begin{figure*}[t]
    \centering
    \includegraphics[width=\linewidth]{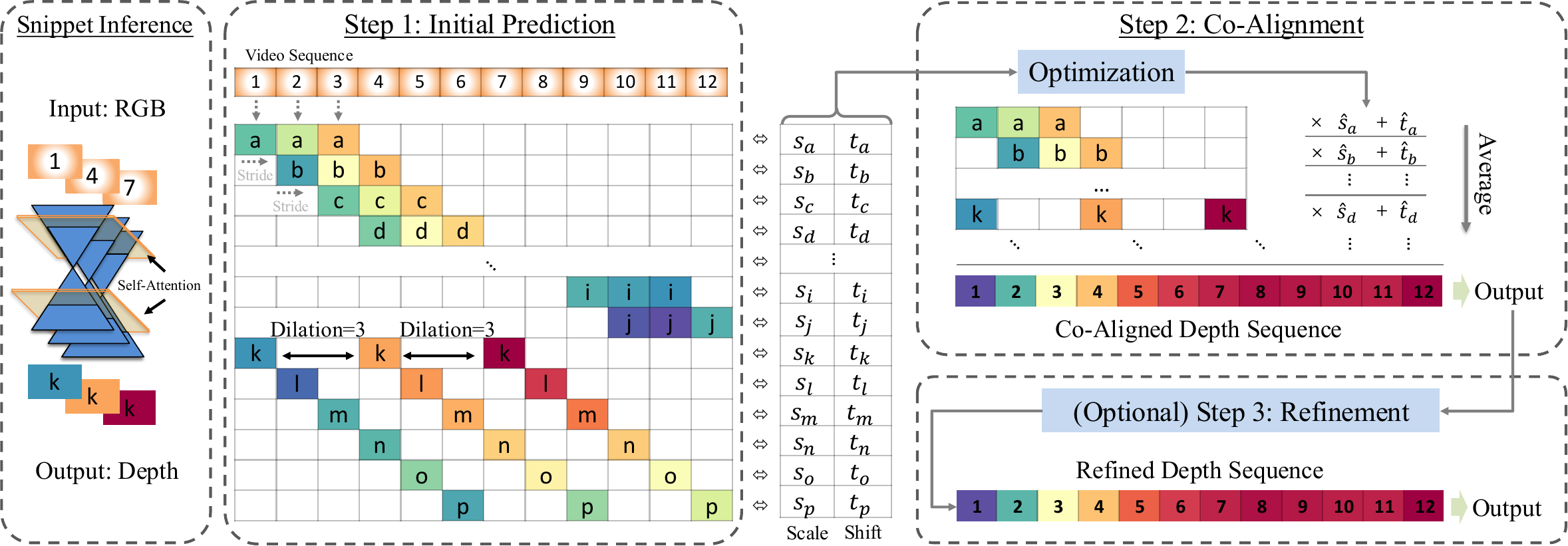}
    \caption{\textbf{Overview of the \method{} Inference Pipeline.} Given a video sequence $\video$ (with~\img~is $i^\text{th}$ frame), we construct $N_T$ overlapping snippets using a \textit{dilated rolling kernel} with varying dilation rates, and perform 1-step inference to obtain initial depth snippets (~\depthSnippet~). Next, depth co-alignment optimizes $N_T$ pairs of scale and shift values to achieve globally consistent depth throughout the full video. An optional refinement step further enhances details by applying additional, snippet-based denoising steps.}
    \label{fig:rolling_kernel}
\end{figure*}

\section{Method} \label{sec:method}
Let $\video \in \mathbb{R}^{N_F \times 3 \times H \times W }$ be an RGB video of length $N_F$, the goal of a monocular video depth estimator is to predict a depth video $\depth \in \mathbb{R}^{N_F \times H \times W }$. All frames in that depth video should share a common depth scale and shift, i.e., depth values should not drift unless the associated pixel moves relative to the camera.
In the following, we present our \method{} framework for predicting $\depth$ from $\video$. The proposed approach is based on a per-snippet LDM, test-time depth co-alignment, and an optional refinement of the resulting video, as illustrated in \cref{fig:rolling_kernel}.

\subsection{Marigold Monocular Depth Recap}
Several recent methods~\cite{ke2023repurposing,duan2023diffusiondepth,saxena2023depthgen}, including our base model Marigold~\cite{ke2023repurposing}, cast monocular depth estimation as conditional image generation, where a pre-trained LDM is retargeted to generate the depth map given the input image. To that end, the model progressively adds noise to depth samples $\depth^i$ and learns to reverse that degradation, to approximate the conditional distribution $\prob(\depth^i|\video^i)$.

In detail, the model is trained to predict the added noise $\epsilon$
at each step by minimizing the objective
\begin{equation*}
    \mathcal{L}(\theta) = \mathbb{E}_{(\depth^i_0,\video^i)\sim\Prob_{\depth^i,\video^i},t\sim\mathcal{U},  \epsilon\sim\mathcal{N}} \left[ \left\| \mathbf{\epsilon} - \mathbf{\epsilon}_\theta(\depth^i_t,\video^i, t) \right\|^2 \right].
\end{equation*}
At inference time the model starts from the input $\video^i$ and pure Gaussian noise $\depth^i_T \sim \mathcal{N}(0, \I )$, 
and gradually maps the latter to a depth map $\depth^i_0$ by iteratively applying the learned denoising step. For computational efficiency, the denoising process operates a low-dimensional latent space $\mathcal{Z}$, with an auto-encoder to map images to latent embeddings, and depth maps back to image space~\cite{rombach2022high}.

%%%%%%%%%%%%%%%%%%

\subsection{Extension to Snippets}
Inspired by multi-view diffusion models~\cite{kong2024eschernet,liu2023syncdreamer}, we extend Marigold~\cite{ke2023repurposing} to handle multiple frames by modifying its self-attention layers. In each self-attention block, we flatten tokens from all frames in a snippet into a single sequence, such that the attention mechanism operates across frames and captures spatial and temporal interactions. Unlike video diffusion models with factorized spatial-temporal attention, this approach can handle frames with varying temporal spacing, which makes it possible to sample snippets at lower frame rates and capture long-range dependencies, an advantage when processing long videos.

The original Marigold model predicts (affine-invariant) depth between image-specific near and far planes. This parametrization poses problems for video depth estimation, where the depth range can vary over time. We therefore retrain Marigold to predict inverse depth (like several other monodepth estimators~\cite{Ranftl2020_midas,yang2024depthv2}), which is less sensitive to such variations, particularly in the far field.

\subsection{From Snippets to Video} \label{subsec: inference}
Our multi-frame depth estimator operates on short snippets of $n$ frames, where $n \ll N_F$. As these snippets are processed independently, each has its own scale and shift -- which are arbitrary in the case of affine-invariant methods~\cite{Ranftl2020_midas,ke2023repurposing,gui2024depthfm} including Marigold, but will in practice not be perfectly aligned even when using a metric depth estimator~\cite{yin2023metric3d,piccinelli2024unidepth,li2024patchrefiner}.
To resolve that ambiguity, we construct overlapping snippets with different temporal dilation rates. The frames shared between different snippets are subsequently used to align all depth predictions to one common scale and shift.

\noindent\textbf{Dilated Rolling Kernel.}
We construct multi-scale snippets using the \textit{dilated rolling kernel}. For instance, for 3-frame snippets with dilation rate (frame spacing) $g$ and stride $h$, the kernel picks frames $(\video^{i-g}, \video^i, \video^{i+g})$ from the input video, where $i\in\{g+\!1, g\!+\!1\!+\!h, g\!+\!1\!+\!2h,\hdots\}$. By varying the dilation rate, we sample snippets with different frame rates, in order to capture temporal dependencies at different time scales. 
For each snippet of $n$ frames, we then predict depth using the multi-frame LDM, to obtain a corresponding $n$-frame depth snippet. 

\noindent\textbf{Depth Co-alignment.}
At this stage we have generated $N_T$ depth snippets. Each of them has its own scale and shift parameters \{$(s_k,t_k)$, $k\in 1\hdots T\}$, which are shared across its constituent frames. Our goal is to jointly compute $N_T$ scale and shift values such that they optimally align all snippets into a consistent video. At a given frame $\video^i$, there are $N^i$ different individual depth maps $\{\depth^i_j,j=1\hdots N^i\}$ originating from different snippets, where $N^i$ can vary from frame to frame. Let $k(i,j)$ be an indexing function that retrieves the snippet index $k$ for the $j$-th depthmap at frame $i$. To estimate the best alignment, we minimize the $L_1$ loss over all individual depth predictions,
\begin{equation}
\label{eq:reg_loss}
    \min_{s_k>0,t_k}
    %\mathcal{L}=
    \left(
    \sum_{i=1}^{N_F} \sum_{j=1}^{N^i}\left|s_{k(i,j)}\depth_j^i + t_{k(i,j)} - \overline{\depth^i} \right|\; 
    \right),
\end{equation}
with the mean depth
\begin{equation}
    \overline{\depth^i} = \frac{1}{N^i}\sum_{j=1}^{N^i} \left(s_{k(i,j)}\depth_j^i + t_{k(i,j)}\right)\;.
\end{equation}
The solution to eq.~\eqref{eq:reg_loss} is found with gradient descent, stabilized by putting more emphasis on snippets with high dilation rates, and additional regularization. Once the depth snippets have been aligned in a common frame with the estimated scale and shift values, the depth maps at every frame $\video^i$ are obtained by taking the pixel-wise mean $\overline{\depth^i}$, resulting in a single, consistent depth video with one depth map per frame. See 
% the supplementary material
Sec.~\ref{sec:supp_coalign} 
for further details.

\begin{figure}
    \centering
    \includegraphics[width=0.9\linewidth]{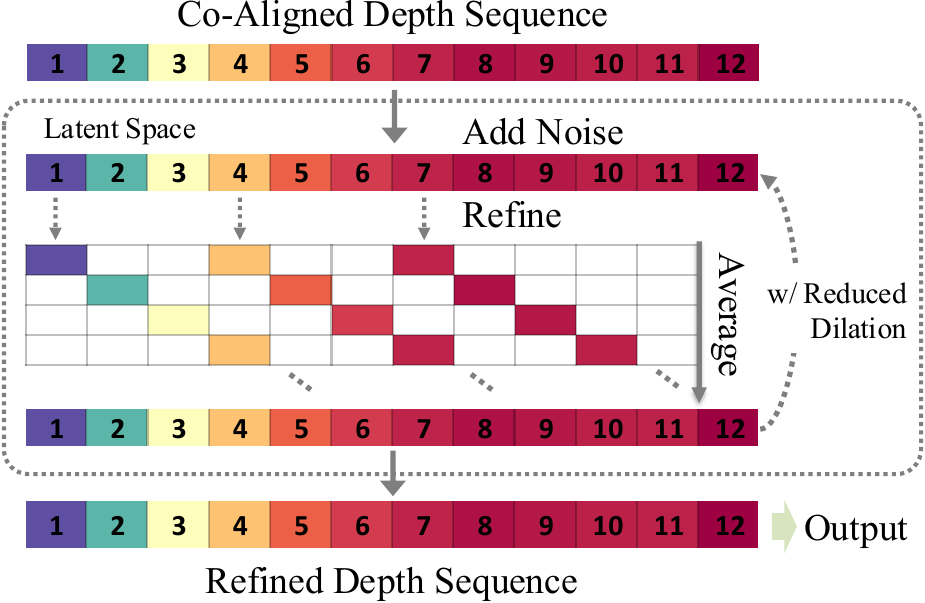}
    \caption{\textbf{Depth Refinement} encodes the co-aligned depth video into latent space, contaminates it with a moderate amount of noise, then denoises it with a series of reverse diffusion steps with decreasing snippet dilation rate. After each step, overlapping latents are averaged to propagate information between snippets.}
    \label{fig:refinement}
\end{figure}

\noindent\textbf{Depth Refinement.}
To enhance visual quality and capture finer details, we optionally apply a diffusion-based refinement step to the merged depth video $\depth$, as illustrated in \cref{fig:refinement}. The video is again encoded into latent space frame by frame, and contaminated with a moderate amount of noise, corresponding to step $T/2$ of the diffusion schedule, halfway between the clean latent and pure noise.
The degraded video is again split into snippets with the dilated rolling kernel and each snippet is denoised individually with the same LDM as above.
To integrate information across overlapping snippets, the latent embeddings
of every frame are averaged after every denoising step. We find that this partial (reverse) diffusion works best when applied in a coarse-to-fine manner in time, starting with a large snippet dilation rate and gradually decreasing it along the denoising process.  
The refinement process enhances high-frequency detail without altering the global scene depth layout, at the cost of increased inference time due to the additional round of denoising diffusion.

\subsection{Multi-Frame Training} \label{subsec:training}

We exploit the flexible design of the multi-frame self-attention mechanism to fine-tune the model with varying snippet lengths. Training snippets are randomly picked to have one, two, or three frames, making sure that the motion between frames is small enough to have overlapping view frustra.
To fully utilize the value range of the diffusion model for best performance, inverse depth values are normalized on a per-snippet basis, using the  2\textsuperscript{nd} and 98\textsuperscript{th} percentiles for robustness.
We found it important to jointly normalize the values within each snippet rather than normalizing each frame individually.
In this way,  the same frame is normalized differently depending on the context it appears in, and normalized depths remain comparable within a snippet, enabling the model to understand and correctly handle rapid changes in the depth range, which routinely appear in longer video sequences.

\begin{table*}[t]
    \centering
    \caption{\textbf{Quantitative comparison} of \method{} with baseline methods on zero-shot benchmarks. \textbf{Bold} numbers are the best, \underline{underscored} second best, numbers in the bracket after each dataset denote video sequence length. \method{} demonstrates superior performance across both short and long video sequences, despite being an image-based model. }
    \vspace{-0.5em}
    \label{tab:main_result}
    \resizebox{\linewidth}{!}{

\begin{tabular}{clcccccccccccccc}
% \begin{tabular}{clllllllllllllll}
\toprule

\multicolumn{1}{l}{} & &
\multicolumn{2}{c}{PointOdyssey (250)} & 
\multicolumn{1}{c}{} & \multicolumn{2}{c}{ScanNet (90)} &
\multicolumn{1}{c}{} & \multicolumn{2}{c}{Bonn (110)} & 
\multicolumn{1}{c}{} & \multicolumn{2}{c}{DyDToF (200)} & 
\multicolumn{1}{c}{} & \multicolumn{2}{c}{DyDToF (100)} \\

\cmidrule{3-4} \cmidrule{6-7} \cmidrule{9-10} \cmidrule{12-13} \cmidrule{15-16} 

\multicolumn{1}{l}{} &  & \multicolumn{1}{c}{Abs Rel↓} & \multicolumn{1}{c}{$\delta1 $ ↑} & \multicolumn{1}{c}{} & \multicolumn{1}{c}{Abs Rel↓} & \multicolumn{1}{c}{$\delta1 $ ↑} & \multicolumn{1}{c}{} & \multicolumn{1}{c}{Abs Rel↓} & \multicolumn{1}{c}{$\delta1 $ ↑} & \multicolumn{1}{c}{} & \multicolumn{1}{c}{Abs Rel↓} & \multicolumn{1}{c}{$\delta1 $ ↑} & \multicolumn{1}{c}{} & \multicolumn{1}{c}{Abs Rel↓} & \multicolumn{1}{c}{$\delta1 $ ↑} \\ 

\midrule

\multirow{3}{*}{\rotatebox{90}{
\begin{tabular}[c]{c}
Single\\frame
\end{tabular}
}}
& Marigold$^*$~\cite{ke2023repurposing} & 
14.9 &   80.4 &&
14.9 &   78.3 &&
10.5 &  86.7 &&
25.3 &  55.5 &&
16.4 & 73.5 \\

& DepthAnything~\cite{yang2024depthv1} & 
16.3 &   76.0 && 
12.9 &   84.0 && 
9.9  &  89.4 &&  
25.4 &  54.3 &&  
16.4 & 75.6 \\

& DepthAnythingv2~\cite{yang2024depthv2} & 
14.4 &   81.4 && 
13.3 &   82.6 &&
10.5 &  87.4 && 
24.8 &  55.9 &&  
16.0 & 76.6 \\

\midrule

\multirow{6}{*}{\rotatebox{90}{Video}}
& NVDS (DPT-Large)~\cite{NVDS} & 
26.6 &   68.2 &&
18.5 &   67.7 &&
10.5 &  88.1 && 
24.7 &  56.0 &&  
18.8 & 69.3 \\

& ChronoDepth~\cite{shao2024learning} &
51.7 &   71.2 &&
16.8 &   73.8 && 
10.9 &  86.9 &&  
26.9 &  53.2 &&  
19.9 & 66.5 \\

& DepthCrafter~\cite{hu2024depthcrafter} & 
36.3 &   75.0 && 
12.7 &   84.3 &&  
\textbf{6.6}  &  \textbf{96.7} && 
22.1 &  60.7 && 
16.2 & 74.7 \\

\cmidrule{2-16} 

& \method~(ours, fast)$^{\dag}$ &
\textbf{9.6}  &  \underline{90.4}  && 
\underline{10.1} &  \underline{89.7}  &&
\underline{7.9}  &  93.6 &&  
\underline{17.7} &  \underline{69.6} && 
\underline{12.7} & \underline{81.6} \\

% & Ours (main) &
% 9.5	 & 90.6	&&
% 9.3	 & 91.4	&&
% 7.8	 & 93.6	&&
% 17.3 & 71.2	&& 
% 12.3 & 83.0 \\

& \method~(ours) & 
\textbf{9.6}  &  \textbf{90.5}  && 
\textbf{9.3}  & \textbf{91.6}  &&  
\underline{7.9}  &  \underline{93.9} &&  
\textbf{17.3} &  \textbf{71.7} && 
\textbf{12.3} & \textbf{83.0} \\

\bottomrule
\end{tabular}%
    }
    \begin{minipage}{0.96\linewidth}
        \scriptsize
        \vspace{0.4em}
        \begin{itemize}
        \item[] $^*$Inverse depth version, retrained with the original training code.
        $^{\dag}$Run at half-precision (fp16), with dilation rates \{1, 25\}, without refinement.
        \end{itemize}
    \end{minipage}
\end{table*}

\begin{figure*}[ht]
    \centering
    \includegraphics[width=1\linewidth]{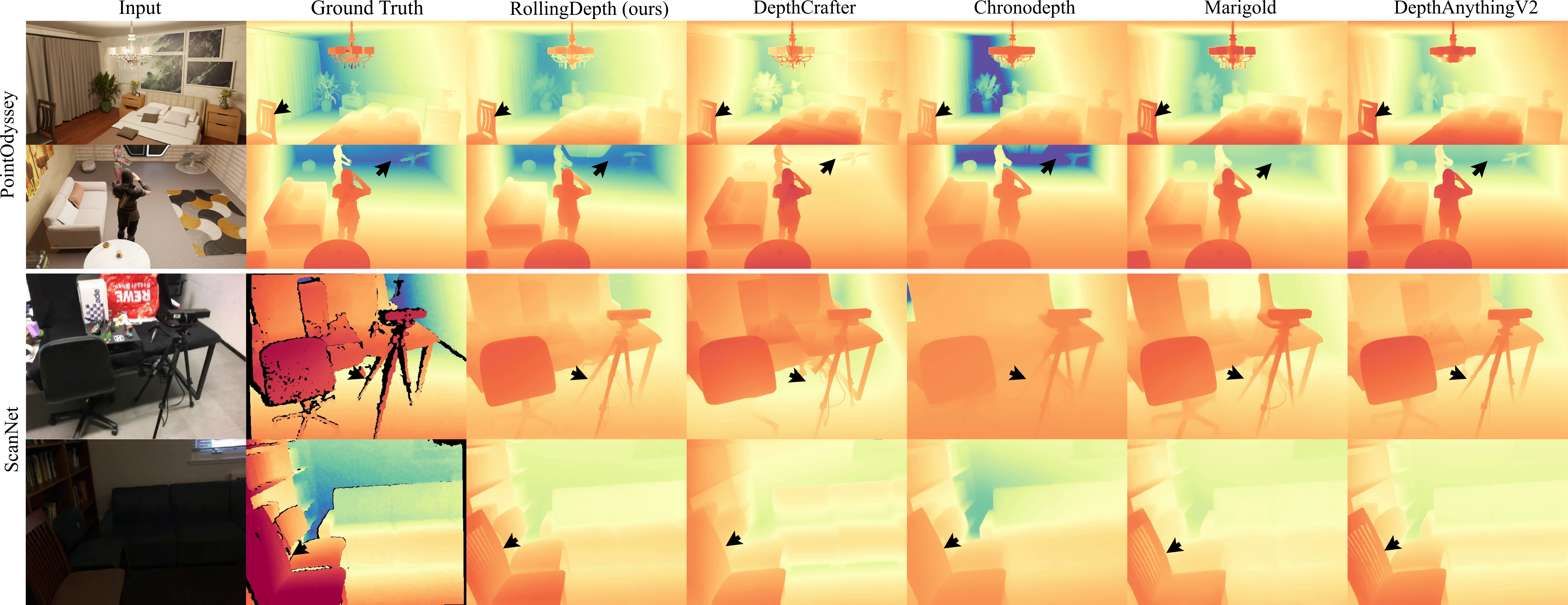}
    \vspace{-1.5em}
    \caption{\textbf{Qualitative comparison} between different methods. \method{} excels at preserving fine-grained details (\cf the chandelier in the first sample and the tripod in the third sample) and recovering accurate scene layout (\cf the far plane in the second sample).}
    \label{fig:qualitative_datasets}
\end{figure*}
%%%%%%%%%%%%%%%%%%%%%%%% Experiment %%%%%%%%%%%%%%%%%%%%%%%%
\section{Experiments}\label{sec:experiment}

\subsection{Implementation Details}

\noindent\textbf{Training Datasets.}
To finetune the snippet LDM, we use TartanAir~\cite{wang2020tartanair}, a synthetic video dataset with various (indoor and outdoor) scenes, styles, and camera motions. We visually inspect the scenes and select 18 scenes consisting of 369 sequences.
Training snippets are randomly sampled from a sequence, with a minimum overlap ratio of 30\%. To increase the diversity of scenes and avoid a significant sim-to-real gap we additionally use Hypersim~\cite{roberts2021hypersim}, a photorealistic single-image dataset containing 365 diverse scenes, treating images as 1-frame snippets.

\noindent\textbf{Training Settings.}
Training images are resized to $480 \times 640$ for efficiency, with random horizontal flipping as data augmentation. To align with the refinement setting, we employ depth range augmentation, where we randomly squeeze the normalized depth snippets to a smaller range and then slightly rescale and shift the depth range in each frame. 
As optimizer, we use AdamW~\cite{loshchilov2019adamw} with a learning rate of $3 \times 10^{-5}$ and exponential decay. Training is run on four Nvidia A100 GPUs with a batch size of 32 and takes approximately 18k iterations or two days to converge.

\noindent\textbf{Inference Settings.}
During inference, we fix the snippet length to $n\!\!=\!\!3$, with three different dilation rates $g \in \{1, 10, 25\}$ to capture short- to mid-range temporal relations. For each snippet we perform 1-step inference.   
Long-range temporal relations are covered by the depth co-alignment, which is initialized with $s_k\!\!=\!\!1$ and $t_k\!\!=\!\!0$ and optimized with 2000 steps of gradient descent, using the Adam optimizer.
For the optional refinement, we start at timestep $T/2$ of the diffusion trajectory and perform 10 denoising steps, gradually reducing the dilation rate from 6 to 1.
Input images are resized to a maximum side length of 768 pixels. For evaluation, the final result is up-sampled to match the original resolution in the dataset.

\subsection{Evaluation}
\noindent\textbf{Evaluation Datasets.}
We evaluate \method{} on four datasets that include both static and dynamic scenes with varying camera and scene motions:
\textbf{PointOdyssey}~\cite{zheng2023pointodyssey} is a synthetic dataset with individually animated characters that move independently, designed for long-term tracking. We filter out overly simplistic toy scenes and retain 35 sequences. For each sequence, we follow the videodepth literature and exclude frames with camera zoom, then select the first 250 frames in each sequence.
\textbf{ScanNet v2}~\cite{Dai_2017_CVPR_ScanNet} is an indoor dataset of static scenes recorded with the Kinect RGB-D sensor. We use its test set of 100 sequences, taking the first 270 frames of each sequence and downsample the frame rate by a factor 3, 
\textbf{Bonn RGBD}~\cite{palazzolo2019refusion} is an RGB-D dataset of moving people in indoor spaces.
% captured with the OptiTrack Prime13 sensor. 
Following \cite{hu2024depthcrafter}, we use frames 30-140 from five different dynamic scenes.
\textbf{DyDToF}~\cite{sun2022consistent} is a photorealistic synthetic dataset featuring moving objects including people and animals. It has several videos per scene, we always take the first video and create two subsets of different lengths from it, by clipping frames 50-250, respectively frames 50-150.

\noindent\textbf{Evaluation Protocol.}
We extend the affine-invariant depth evaluation protocol~\cite{Ranftl2020_midas} to videos, i.e., depth predictions $\hat{\depth}$ are aligned to the ground truth with a scale and shift found with least squares fitting, where we fit one pair of transformation parameters per video, i.e., all frames in a video share a common scale and shift.
We quantify the depth estimation accuracy with two standard metrics~\cite{Ranftl2020_midas, ke2023repurposing, NVDS, hu2024depthcrafter}: the \emph{absolute mean relative error} (\textit{AbsRel}), defined as $\frac{1}{M} \sum_{j=1}^M {|\hat{\depth_j} - \depth_j|} / {\depth_j}$, where $M$ is the total number of pixels; and the $\delta1$-accuracy, which measures the fraction of pixels for which $\max(\hat{\depth_j}/\depth_j, \depth_j/\hat{\depth_j}) < 1.25$.
Metrics are always given as percentages. We provide additional temporal smoothness evaluation in the supplementary material.

\begin{figure*}[t]
    \centering
    \includegraphics[width=\linewidth]{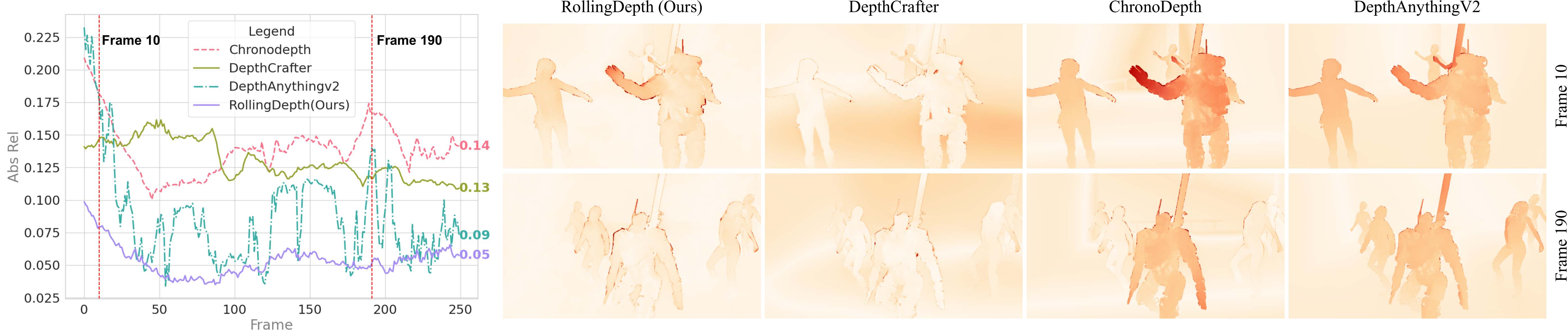}
    \vspace{-1.5em}
    \caption{\textbf{AbsRel error over time:}  The line plot (\textit{left}) shows the depth error at every individual frame, end-of-line numbers are the average error across the video. The images (\textit{right}) display error maps (low~\orrd~high) for two specific frames. \method{}~achieves the lowest error overall, competing methods recover scene layout less faithfully and tend to be biased towards the foreground or the background. }
    \label{fig:qualitative_diff}
\end{figure*}

\begin{figure*}[t]
    \centering
    \includegraphics[width=1\linewidth]{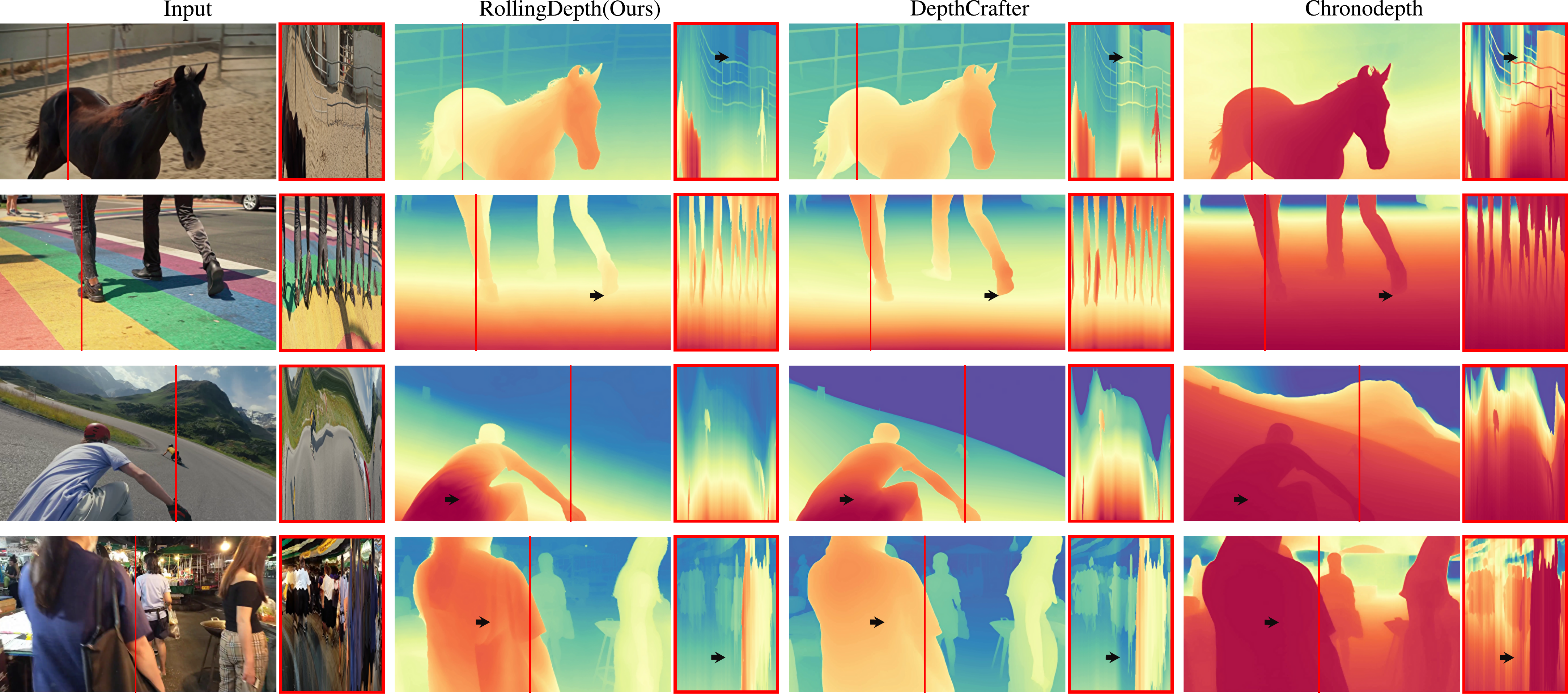}
    \vspace{-1.5em}
    \caption{\textbf{Qualitative comparison} of depth predictions (near~\spectral~far) from in-the-wild videos. To graphically show temporal consistency, we display temporal profiles (\textit{red box}) for a fixed column (marked with a red line). \method{} picks up subtle details like accessories and wrinkled cloth, and mitigates spurious depth discontinuities (\cf background in temporal profile of the first sample) in time.}
    \label{fig:qualitative_inthewild}
\end{figure*}

\subsection{Comparison with Other Methods}
We compare \method{} against six state-of-the-art methods for zero-shot monocular depth estimation: 
Marigold~\cite{ke2023repurposing}, DepthAnything~\cite{yang2024depthv1} and DepthAnythingv2~\cite{yang2024depthv2}, which are single-frame methods; as well as NVDS~\cite{NVDSPLUS}, ChronoDepth~\cite{shao2024learning}, and DepthCrafter~\cite{hu2024depthcrafter}, which are video-based approaches.

\noindent\textbf{Quantitative Comparison.}
As shown in~\cref{tab:main_result}, \method{}  outperforms both single-frame and video-based approaches across multiple datasets and different sequence lengths, often by considerable margins.
We attribute this to its ability to combine the accuracy of image-based models with the temporal coherence afforded by our snippet-based inference and global depth co-alignment.
On PointOdyssey, which includes many challenging scenes with highly variable depth ranges, \method{} achieves by far the best result. Methods based on video models struggle on this dataset, and are in fact even unable to match the performance of single-frame methods. 
We observe that the performance of video models drops especially in scenes with sudden, large changes in the depth range (\eg a hand gesture in front of the camera). We hypothesize that the underlying video prior is too rigid and prevents a correct adaptation to the rapid change, see 
% the supplementary material 
Sec.~\ref{sec:supp_video_model_failure} 
for details.
Also on DyDToF, \method{} greatly reduces the error compared to other methods, again underscoring its ability to handle dynamic scenes and variations of the near and far planes.
Still, the good performance is not limited to dynamic scenes with strong depth variations. \method{} also performs well on indoor data, reaching the lowest error on the static ScanNet scenes and the second-lowest error on the Bonn data. Here DepthCrafter shines -- we observe that it generally tends to do well in scenes dominated by foreground objects, particularly humans.

\noindent\textbf{Qualitative Comparison.}
To make our findings more tangible, we provide qualitative comparisons both on evaluation data and on in-the-wild examples.
Figure.~\ref{fig:qualitative_datasets} confirms that \method{} consistently produces high-quality depth maps that preserve fine detail, both near the camera in the distance.
DepthCrafter and ChronoDepth produce locally smooth videos with little frame-to-frame flicker, but have a tendency to distort the overall scene layout in a way that certain objects are segmented well but placed at incorrect (relative) depths.
Single-image estimators are seemingly more accurate in that respect, but suffer from flickering and a lack of temporal coherence.
We further illustrate these trends in \cref{fig:qualitative_diff}, where we plot per-frame errors, as well as per-pixel errors for selected frames.

To demonstrate generalization to real-world video clips, \cref{fig:qualitative_inthewild} shows depth predictions for videos collected from the internet.
Also in these cases, \method{} accurately recovers fine details and maintains long-term coherence.
To better illustrate the evolution of the depth estimate over time, we extract temporal profiles for fixed image columns. They exhibit no significant high-frequency variations along the time axis that would indicate frame-to-frame flicker.
We also do not observe drift or unwarranted jumps in the depth values that would indicate systematic biases.
DepthCrafter for the most part also recovers plausible depth, but misses depth variations within the main segments and sometimes exhibits instabilities along the time axis.
Chronodepth recovers depth boundaries rather well, but delivers billboard-like, layered depth maps.

\subsection{Ablation Studies}
We validate our main hyper-parameters and design choices on a subset of 10 sequences from the PointOdyssey test set and 20 sequences from the ScanNet test set.

\noindent\textbf{Dilation of Initial Predictions.}
We start by ablating the arguably most crucial hyper-parameter of \method{}, the dilation rate for snippet sampling, see \cref{tab:ablate_gap}.
The base setting uses only dilation rate \{1\} for minimal information exchange and smoothness between adjacent frames. Having a high dilation rate \{1,25\} gives the model access to longer-term motion patterns on the order of 1 second and greatly stabilizes the co-alignment step, which in turn reduces the AbsRel error by \textgreater6 percept points on PointOdyssey and by \textgreater2 percent points on the (static) ScanNet.
This is what we use in our \textit{fast} setting (c.f. \cref{tab:main_result}), which takes 81s for a 768$\times$432 video of 250 frames (ChronoDepth: 121s, DepthCrafter: 284s).
An additional, intermediate dilation rate \{1,10,25\} further intensifies the information exchange across time. This further boosts the quality of the estimated depth maps, but as expected yields diminishing returns.
\begin{table}[t]
    \centering
    \caption{\textbf{Ablation of dilation rates} for snippet prediction. We report values before the optional refinement step. The minimal base setting uses only dilation rate 1. Adding a high dilation rate 25 brings a marked performance gain. Yet another dilation rate 10 gives a further, smaller boost.}
    \label{tab:ablate_gap}
    \resizebox{0.8\linewidth}{!}{%

% \begin{tabular}{lccccc}
% \toprule
%  Dilation(s) of  & \multicolumn{2}{c}{PointOdyssey} & & \multicolumn{2}{c}{ScanNet} \\
%  \cline{2-3} \cline{5-6}
%  Initial Predictions & {Abs Rel↓} & {$\delta1 $ ↑} & & {Abs Rel↓} & {$\delta1 $ ↑} \\
% \hline
% \{0\} &
% 16.7 & 75.5	&& 12.8 & 83.2 \\

% \{0, 25\} &
% \textbf{10.2} & 89.5	&& 10.6 & 88.8 \\

% \{0, 10, 25\} &
% \textbf{10.2} & \textbf{89.6}	&& \textbf{9.9}  & \textbf{90.1} \\

% \bottomrule
% \end{tabular}%

\begin{tabular}{lccccc}
\toprule
  & \multicolumn{2}{c}{PointOdyssey} & & \multicolumn{2}{c}{ScanNet} \\
 \cline{2-3} \cline{5-6}
Dilation rates  & {Abs Rel↓} & {$\delta1 $ ↑} & & {Abs Rel↓} & {$\delta1 $ ↑} \\
\hline
\{1\} &
16.7 & 75.5	&& 12.8 & 83.2 \\

\{1, 25\} &
\textbf{10.2} & 89.5	&& 10.6 & 88.8 \\

\{1, 10, 25\} &
\textbf{10.2} & \textbf{89.6}	&& \textbf{9.9}  & \textbf{90.1} \\

\bottomrule
\end{tabular}%

    }
\end{table}

\noindent\textbf{Effectiveness of Co-Alignment and Refinement.}
We further isolate the effect of the \method{}'s components, see \cref{tab:ablate_component}. The snippet diffusion step is mandatory to obtain any depth estimates at all and cannot be left out. For the experiment we switch on and off the two remaining steps, co-alignment and refinement, and test all combinations.
Simply merging overlapping latents without prior alignment proves to be insufficient, i.e., their individually estimated depth ranges are too inconsistent to average them into a coherent sequence. The refinement step cannot fix that problem.
Conversely, the co-alignment does the heavy lifting to fuse depth snippets with different scales and shifts into a coherent video and contributes the lion's share of the improvement.
Subsequent refinement of the aligned video only results in a marginal increase of the performance metrics, but visibly improves the result by recovering sharp details that have been missed or blurred in the preceding steps.

\begin{table}[t]
    \centering
    \caption{\textbf{Ablation of components.} Depth co-alignment is a crucial functionality for the snippet-based strategy of \method{}, whereas the additional refinement has only a small effect on the performance metrics, despite visibly enhanced image detail.}
    % Refinement setting: 10-step, coarse-to-fine gap
    \label{tab:ablate_component}
    \resizebox{\linewidth}{!}{%

\begin{tabular}{ccccccc}
\toprule
 & & \multicolumn{2}{c}{PointOdyssey}  & & \multicolumn{2}{c}{ScanNet}\\
 \cline{3-4} \cline{6-7}
Co-Alignment & Refinement & {Abs Rel↓} & {$\delta1 $ ↑} & & {Abs Rel↓} & {$\delta1 $ ↑} \\
\hline

$\times$ & $\times$ &
13.0 & 84.4 && 12.4 & 84.3 \\

$\times$ & \checkmark &
13.0 & 84.6 && 12.3 & 84.8 \\

\checkmark & $\times$ &
\textbf{10.2} & 89.6 && 9.9  & 90.1 \\

\checkmark & \checkmark &
\textbf{10.2} & \textbf{89.8} && \textbf{9.8}  & \textbf{90.2} \\

\bottomrule
\end{tabular}%

    }
\end{table}

%%%%%%%%%%%%%%%%%%%%%%%% Conclusion %%%%%%%%%%%%%%%%%%%%%%%%
\section{Conclusion}
\label{sec:conclusion}

We have introduced \method{}, a novel method for monocular video depth estimation that is derived from a single-image (latent) diffusion model.
The core components of our method (i) are a monodepth estimator for short snippets, sampled at various frame rates to capture temporal context at different time scales; (ii) an optimization-based co-alignment procedure that optimally registers all snippets of a video into a common depth range; and (iii) an optional refinement step, again based on the same denoising diffusion scheme for snippets, that enhances fine details in the depth video.
\method{} strikes a favorable balance between accurate per-frame depth prediction and temporal coherence, and can process long video with hundreds of frames.
It empirically delivers best-in-class performance across multiple datasets, also outperforming alternatives derived from full-blown video diffusion models.
That being said, the \method{} framework is flexible and offers the possibility to replace individual components. For instance, an interesting avenue for future work would be to swap out the snippet-based refinement and replace it with a generative video model or a flow-based method for even better motion reconstruction.

%%%%%%%%%%%%%%%%%%%%%%%% Acknowledgment %%%%%%%%%%%%%%%%%%%%%%%%
\noindent \textbf{Acknowledgments.} We thank Yue Pan, Shuchang Liu, Nando Metzger, and Nikolai Kalischek for fruitful discussions. We are grateful to \emph{redmond.ai} for providing GPU resources.

%%%%%%%%%%%%%%%%%%%%%%%% Reference %%%%%%%%%%%%%%%%%%%%%%%%
{
    \small
    \bibliographystyle{ieeenat_fullname}
    \bibliography{main}
}

%%%%%%%%%%%%%%%%%%%%%%%% Supplement %%%%%%%%%%%%%%%%%%%%%%%%
\clearpage
\appendix
\section*{\Large Supplementary Material}

\renewcommand*{\thesection}{\Alph{section}}
\newcommand{\multiref}[2]{\cref{#1}--\ref{#2}}
\renewcommand{\thetable}{S\arabic{table}}
\renewcommand{\thefigure}{S\arabic{figure}}

\setcounter{table}{0}
\setcounter{figure}{0}

This supplementary material includes additional implementation details and experimental results. 
\vspace{-.2cm}

\section{Implementation Details}\label{sec:supp_implement}
\subsection{Depth Co-Alignment}\label{sec:supp_coalign}

As discussed in 
% Sec.~3.3 in the main paper, 
\cref{subsec: inference},
let $k(i,j)$ denote an indexing function that returns the snippet index $k$ corresponding to the $j$-th depthmap of $i$-th frame.
To make the optimization more robust,
we include an additional loss term in depth space while predicting inverse depth.
We further scale the loss terms by their respective mean absolute value per frame to increase the numerical stability.
Additionally, soft constraints on $s_k,t_k$ are applied:
\begin{align}
    \min_{s_k>0,t_k}
    &\left(
    \sum_{i=1}^{N_F} \sum_{j=1}^{N^i}\left|\frac{\widehat{\depth_j^i}  - \overline{\depth^i}}{\widehat{\mu_i}} \right| \right.  + \left. \left|\frac{\widehat{\depth_j^i}^{-1}  - \widetilde{\depth^i}}{\widetilde{\mu_i}} \right|
    \right) \notag \\
    & + \lambda_1 \max(0, 1-s_{k(i,j)})^2 
    + \lambda_2 t_{k(i,j)},
\end{align}
where $\widehat{\depth_j^i}  = s_{k(i,j)}\depth_j^i + t_{k(i,j)}$.
The mean depth and mean inverse depth are defined as
\begin{equation}
    \overline{\depth^i} = \frac{1}{N^i}\sum_{j=1}^{N^i} \widehat{\depth_j^i} \quad \quad 
    \widetilde{\depth^i} = \frac{1}{N^i}\sum_{j=1}^{N^i} \widehat{\depth_j^i}^{-1},
\end{equation}
with the corresponding mean absolute values per frame given by
\vspace{-.35cm}
\begin{equation}
\widehat{\mu_i} =  \frac{1}{HW}\sum_{}^{HW} \left| \overline{\depth^i}\right| \quad \quad 
\widetilde{\mu_i} =  \frac{1}{HW}\sum_{}^{HW} \left| \widetilde{\depth^i}\right|.
\end{equation}
We found that $\lambda_1=10^{-1}$, $\lambda_2=10^{1}$ work well in practice.

\subsection{Additional Training and Inference Details}
During training, we follow Marigold to use MSE loss on the latents. We apply gradient accumulation to increase the effective batch size, to 32.
To better mix the samples with varying snippet lengths, every mini-batch is sampled randomly and can have different snippet lengths.
For the initial depth prediction, we apply the same random Gaussian noise to all frames. When applying refinement, 
the same noise is used to perturb the (encoded) co-aligned depth sequence. The denoising process then starts from timestep $T/2$.

\subsection{Evaluation Datasets}
\begin{figure}[h] 
    \begin{subfigure}[b]{0.45\linewidth}
        \includegraphics[width=\linewidth]{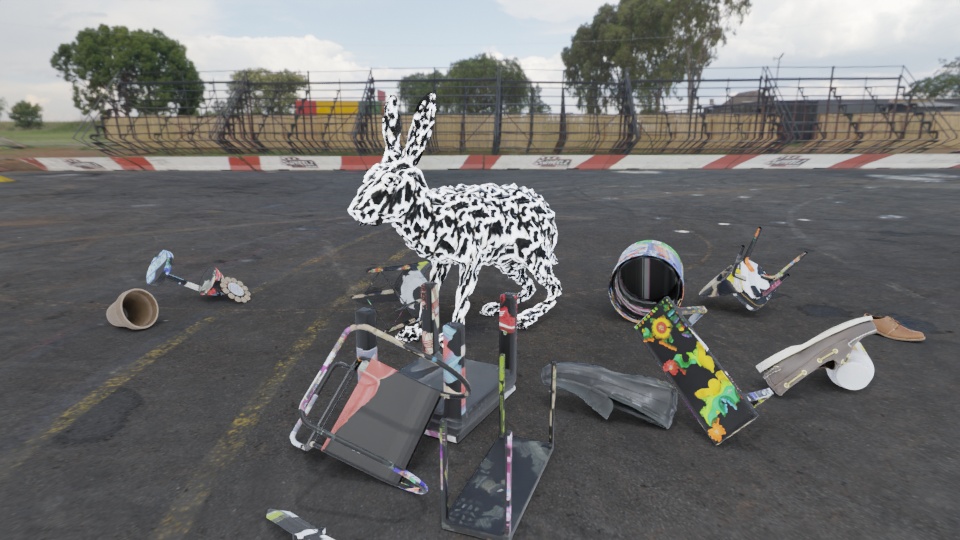}
    \end{subfigure}
    \hfill
    \begin{subfigure}[b]{0.45\linewidth}
        \includegraphics[width=\linewidth]{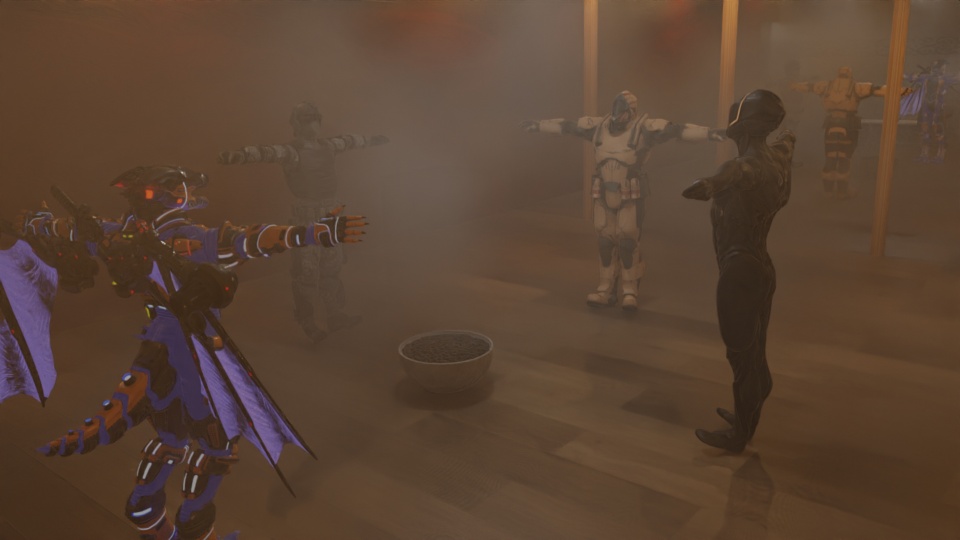}
    \end{subfigure}
    \caption{Examples of PointOdyssey toy scenes (\textit{left}) and scenes with smoke (\textit{right}).}
    \label{fig:supp_po_sample}
\end{figure}

\textbf{PointOdyssey~\cite{zheng2023pointodyssey}} contains several sequences that feature overly simplified toy scenes, as well as some with smoke, for which depth estimation is ambiguous (\cf \cref{fig:supp_po_sample}). We exclude these sequences from the test dataset, a detailed list of selected frames will be provided with the code. For evaluation, pixels on windows are excluded due to inconsistent depth labels.

In \textbf{ScanNet~\cite{dai2017scannet}}, the RGB images and depth labels include a thin black border. Following DepthCrafter~\cite{hu2024depthcrafter}, we crop the RGB images by removing 8 pixels from the top and bottom and 12 pixels from the left and right. Similarly, we crop the depth maps by removing 4 pixels from the top and bottom and 6 pixels from the left and right.

For \textbf{DyDToF~\cite{sun2022consistent}}, we exclude depth values beyond 23m, corresponding to less than 1\% of the depth values.

\vspace{-.5cm}

\subsection{Baseline Methods}
We evaluate baseline methods using their recommended default settings. For DepthCrafter~\cite{hu2024depthcrafter}, the inference is performed with 25 diffusion steps, using an overlap of 25 frames for videos longer than 110 frames. For ChronoDepth~\cite{shao2024learning}, inference comprises 10 denoising steps, with a window size of 10 (referred to as ``num-frames" in the code) and a stride of 9 (referred to as ``denoise-steps" in the code).

For Marigold~\cite{ke2023repurposing}, we retrained an inverse depth version using the trailing scheduler setting~\cite{garcia2024e2e, lin2024common}. Under this configuration, 1-step inference with a single model achieves performance comparable to the original configuration with multi-step inference and ensembling, so we utilize the former, more efficient setting.
\vspace{-.1cm}

\section{Additional Experiment Results}\label{sec:supp_exp}

\subsection{Temporal smoothness evaluation}
We further quantitatively evaluate the temporal smoothness using optical-flow-based warping loss (OPW)~\cite{wang2022less} on PointOdyssey and ScanNet datasets and report the results in Tab.~\ref{tab:supp_opw}. The optical flow is estimated using GMFlow~\cite{xu2022gmflow}.

\begin{table}[h!]
\centering
\caption{Temporal smoothness (OPW↓) comparison. All values are $\times 10^3$, lower is better.
$^*$ denotes catastrophic failures on some sequences. Numbers in brackets are evaluated on subsets that exclude those cases. 
} \label{tab:supp_opw}
\resizebox{0.8\linewidth}{!}{
    \begin{tabular}{lcc}
        \toprule
                      & PointOdyssey & ScanNet  \\
        \hline
        Marigold      & 3.52\textcolor{white}{$^*$} (4.00) & 0.48   \\
        DepthAnything & 3.92\textcolor{white}{$^*$} (4.21) & 0.32   \\
        NVDS          & \underline{3.50}\textcolor{white}{$^*$} (2.97) & 0.29 \\
        ChronoDepth         & 8.98$^*$ (2.99)  & 0.29  \\
        DepthCrafter        & 7.75$^*$ (\textbf{1.30}) & \underline{0.25}     \\
        RollingDepth (ours) & \textbf{1.42}\textcolor{white}{$^*$} (\underline{1.63})  & \textbf{0.20} \\
        \bottomrule
    \end{tabular}
}
\end{table}

We notice that ChronoDepth and DepthCrafter have catastrophic failure in some cases of PointOdyssey (\cf Sec.~\ref{sec:supp_video_model_failure}), leading to large errors, as denoted by $^*$. We manually exclude these failure cases. The re-calculated average OPW is reported in the brackets.
Overall, RollingDepth shows good smoothness, on par with DepthCrafter, while being more robust than DepthCrafter and ChronoDepth against occasional failures. 

We point out that OPW only evaluates the ``smoothness” between adjacent frames while ignoring the long-term smoothness and geometric consistency. As shown in Tab.~\ref{tab:supp_dilation}, with larger dilation rates, the geometric accuracy shows a clear improving trend, while the trend of OPW is unclear. We hypothesize that with a larger dilation rate, geometric accuracy is improved at a cost of minor local smoothness decrease when merging the aligned snippets.

\begin{table}[htb!]
    \centering
    \caption{Extended table of dilation rate ablation study (Tab.~2). Values are $\times 10^3$.} \label{tab:supp_dilation}
    \vspace{-1em}
    \resizebox{\linewidth}{!}{
    \begin{tabular}{lccccccc}
        \toprule
          & \multicolumn{3}{c}{PointOdyssey} & & \multicolumn{3}{c}{ScanNet} \\
         \cline{2-4} \cline{6-8}
        Dilation rates  & {Abs Rel↓} & {$\delta1 $ ↑} & \textbf{OPW ↓} & & {Abs Rel↓} & {$\delta1 $ ↑} & \textbf{OPW ↓} \\
        \hline
        \{1\} &
        16.7 & 75.5	& 1.22 && 12.8 & 83.2 & 0.24 \\
        
        \{1, 25\} &
        10.2 & 89.5 & 2.06 && 10.6 & 88.8 & 0.29 \\
        
        \{1, 10, 25\} &
        10.2 & 89.6 & 1.98 && 9.9  & 90.1 & 0.29 \\
        
        \bottomrule
    \end{tabular}
    }
\end{table}

\subsection{Evaluation on DDAD dataset}
We further evaluate the model performance on the DDAD~\cite{packnet} dataset, which is a driving-scene dataset with sparse depth annotation. We use the 100-frame sequences on the test set.

As shown in Tab.~\ref{tab:ddad}, RollingDepth outperforms other methods in terms of accuracy and smoothness.

\begin{table}[h!]
\centering
\centering
\caption{Evaluation on DDAD dataset.} \label{tab:ddad}
\vspace{-1em}
\resizebox{0.75\linewidth}{!}{
\begin{tabular}{lcccccc}
\toprule
    & Abs Rel↓ & $\delta1$ ↑ & OPW ↓ \\
    & {\footnotesize $\times 10^{-2}$} & {\footnotesize $\times 10^{-2}$} & {\footnotesize $\times 10^{-3}$} \\
\hline
NVDS & 30.8 & 57.2 & 0.39 \\
ChronoDepth & 34.2 & 46.9 & \underline{0.21} \\
DepthCrafter & \underline{19.3} & \underline{74.8} & 0.28 \\
RollingDepth (ours) & \textbf{12.8} & \textbf{83.2} & \textbf{0.19} \\
\bottomrule
\end{tabular}
}
\end{table}

\subsection{Inference efficiency}

We report the inference efficiency comparison in Tab~\ref{tab:efficiency}. The benchmarking is done on the same machine with a single RTX3090 GPU. For each method, we run 10 repeated inferences after a warm-up iteration, with the model loaded on GPU, and calculated the mean run time and peak memory footage of each iteration.

\begin{table}[h!]
    \centering
    \caption{Inference speed and peak GPU memory usage comparison on a 768$\times$432 video of 250 frames. By increasing the batch size of processing, $\text{RollingDepth}^\dag$ can trade memory for speed.} \label{tab:efficiency}
    \resizebox{\linewidth}{!}{
    \begin{tabular}{lcccc}
    \toprule
        & Time (s) & Peak GPU Memory (GB) \\
    \hline
    NVDS & 284 & \textcolor{white}{1}\underline{7.6} \\
    ChronoDepth & 121 & 15.0 \\
    DepthCrafter & 284 & 13.6 \\
    RollingDepth (ours) & \underline{105} & \textcolor{white}{1}\textbf{6.2} \\
    $\text{RollingDepth}^\dag$ (ours) & \textcolor{white}{1}\textbf{81} & 40.1 \\
    \bottomrule
    \end{tabular}
    }
\end{table}

\subsection{Failure cases of video models on PointOdyssey}\label{sec:supp_video_model_failure}
\begin{figure*}[htb]
    \centering
    \includegraphics[width=\linewidth]{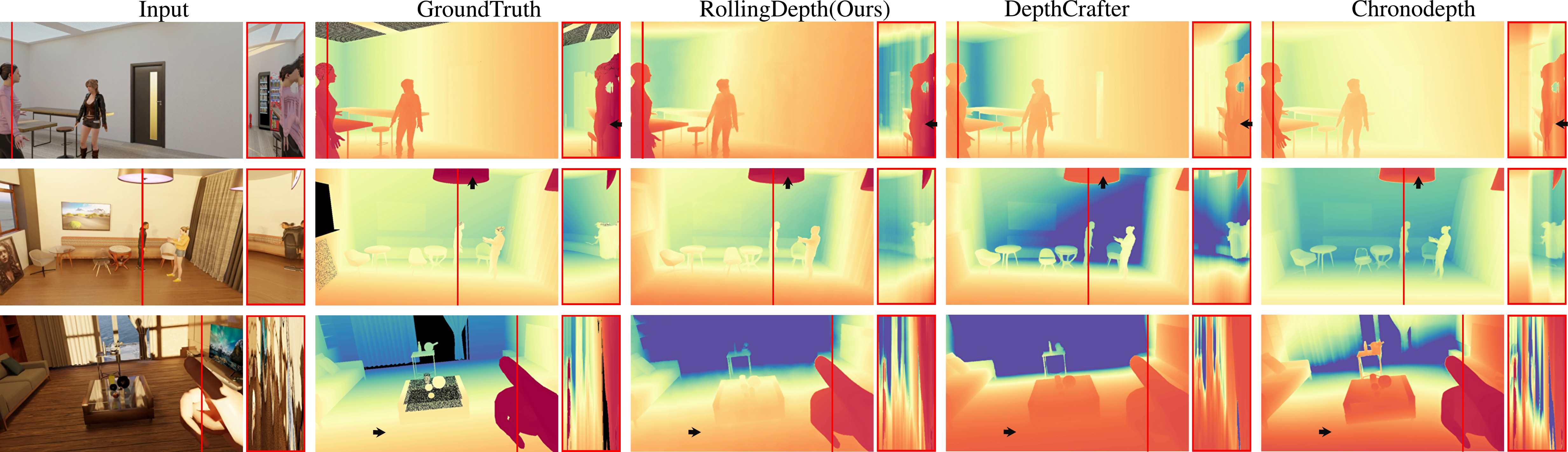}
    \caption{Examples of PointOdyssey samples that challenge video models. In the cases above, the (inverse) depth range varies significantly across frames. The arrows highlight situations where video models yield distorted depth maps.
    In the first two rows, this occurs in regions where the depth deviates significantly from the surrounding scene.
    In the last row, the depth predictions get drawn towards the near plane to match the object close to the camera, biasing the depth in the far field.}
    \label{fig:supp_po_sample_fail}
\end{figure*}

We provide further examples from the PointOdyssey dataset where video-based methods struggle.
\Cref{fig:supp_po_sample_fail} features scenes with large depth changes, such as hand gestures in front of the camera or objects entering the near field.
These sudden changes require rapid alterations of the depth range, both before and after the event. Video models tend to produce incorrect overall scene layout in such cases, we hypothesize that they "try too hard" to equalize the depth range throughout the scene.

\subsection{Failure Cases of \method}
\begin{figure*}[htb]
    \centering
        \includegraphics[width=\linewidth]{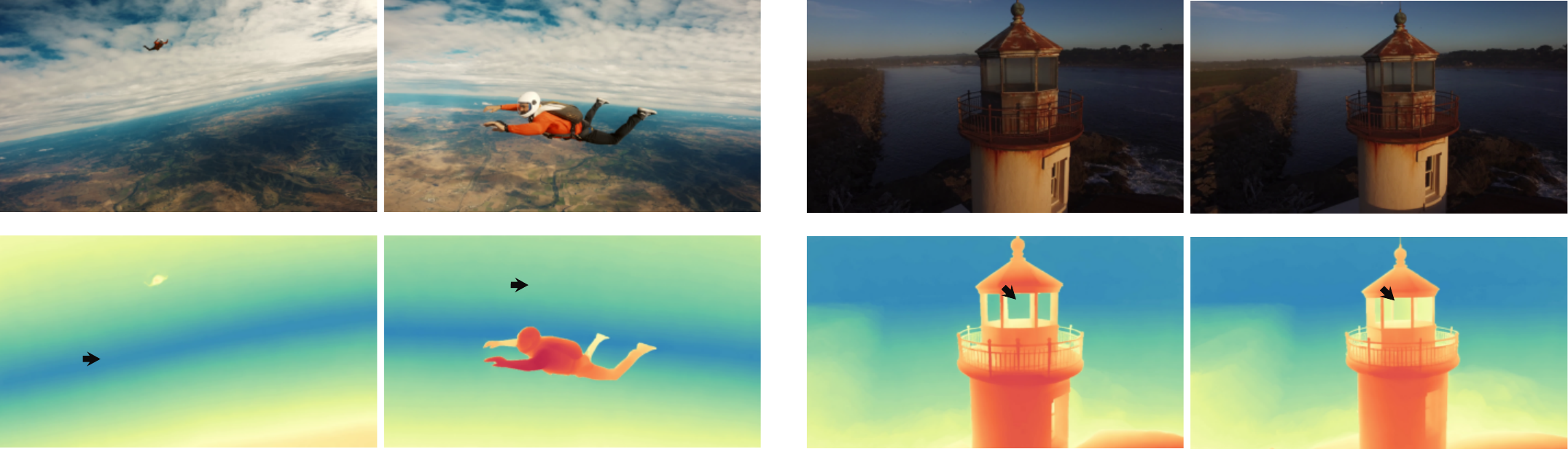}
        \caption{The two samples on the left show incorrect depth predictions in the cloudy sky. 
        The two samples on the right show inconsistencies between different frames of the same video, where the depth at the glass windows fluctuates between the solid and transparent states.}
        \label{fig:fails}
\end{figure*}

While our proposed method handles changing depth range more robustly than video models, it also has certain limitations. Two examples are shown in \cref{fig:fails}. \method{} sometimes misjudges the depth of cloudy skies. Another source of error is transparent surfaces such as glass windows, where subtle variations of transparency or reflections may cause the depth to oscillate between the glass and the scene behind it -- a common issue of depth estimators.

\end{document}